\documentclass[preprint,12pt]{elsarticle}

\usepackage{soul}
\usepackage{amssymb}
\usepackage{amsmath}
\usepackage{comment}
\usepackage{hyperref}
\usepackage{booktabs}
\usepackage{multirow}
\usepackage{tabularx}
\usepackage{textcomp}
\usepackage{lscape}
\usepackage{graphicx}
\usepackage{orcidlink}
 
\journal{Automation in Construction}

\begin{document}

\begin{frontmatter}



\title{Deep Filament Extraction for 3D Concrete Printing}

\author[1]{Karam Mawas\,\orcidlink{0000-0002-8608-7578}}\fnref{cor1}\fnref{equal1}\ead{k.mawas@tu-bs.de}

\author[1]{Mehdi Maboudi \orcidlink{0000-0003-3367-2404}}\fnref{equal1}\ead{m.maboudi@tu-bs.de}

\author[1]{Pedro Achanccaray \orcidlink{0000-0002-7324-9611}}\fnref{equal1}\ead{p.diaz@tu-bs.de}

\author[1]{Markus Gerke \orcidlink{0000-0002-2221-6182}}\fnref{equal1}\ead{m.gerke@tu-bs.de}

\fntext[cor1]{Corresponding author.}


\affiliation[1]{%
  organization={Technische Universität Braunschweig, Institute of Geodesy and Photogrammetry},
  addressline={Bienroder Weg 81},
  city={Braunschweig},
  postcode={38106},
  state={Lower Saxony},
  country={Germany}}
\begin{abstract}
The architecture, engineering and construction (AEC) industry is constantly evolving to meet the demand for sustainable and effective design and construction of the built environment. In the literature, two primary deposition techniques for large-scale 3D concrete printing (3DCP) have been described, namely extrusion-based (Contour Crafting\textemdash~CC) and shotcrete 3D printing (SC3DP) methods. The deposition methods use a digitally controlled nozzle to print material layer by layer. The continuous flow of concrete material used to create the printed structure is called a filament or layer. As these filaments are the essential structure defining the printed object, the filaments’ geometry quality control is crucial. This paper presents an automated procedure for quality control (QC) of filaments in extrusion-based and SC3DP printing methods.  The paper also describes a workflow that is independent of the sensor used for data acquisition, such as a camera, a structured light system (SLS) or a terrestrial laser scanner (TLS). This method can be used with materials in either the fresh or cured state. Thus, it can be used for online and post-printing QC.
\end{abstract}


\begin{graphicalabstract}
\begin{figure}
\centering
\includegraphics[width=1\textwidth]{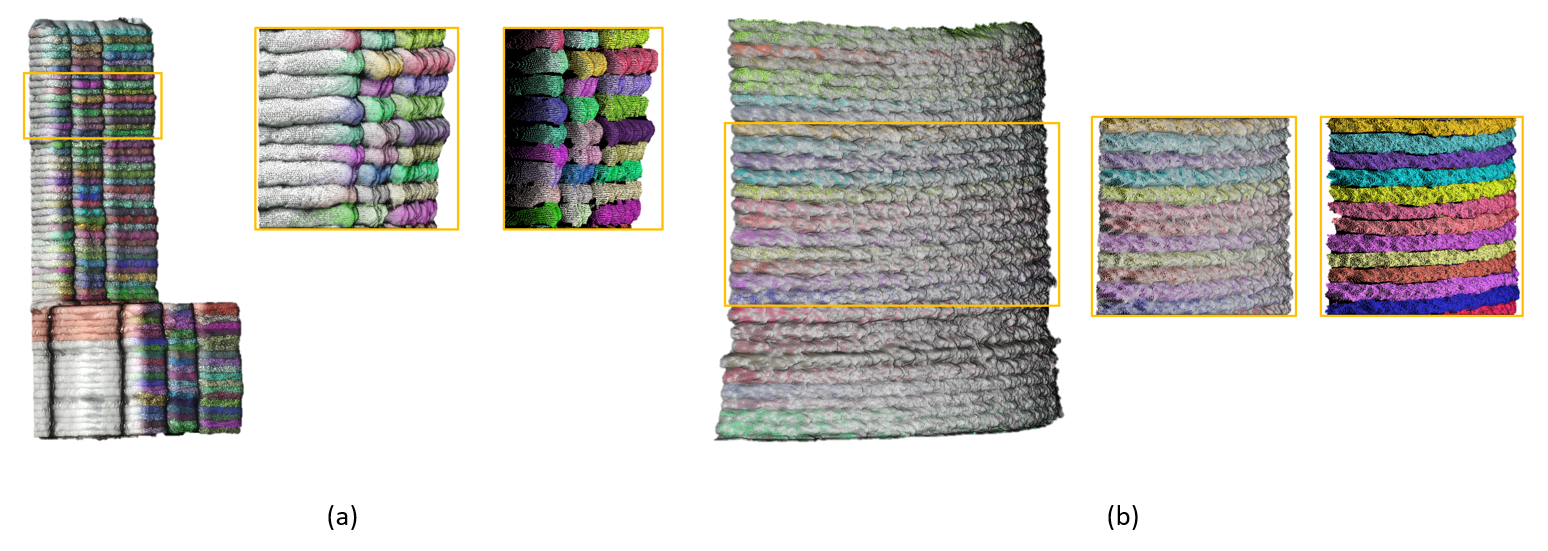}
\caption{The prediction results of the filament instances are projected back into 3D space. (a) Extrusion-based object from Photogrammetry: The results shown are for the data from Fig.~\ref{Fig: Results_imposed}b. (b) SC3DP object from TLS: The results shown for the data from Fig.~\ref{Fig: Results_imposed}c. The left image shows the raw data of the whole object superimposed with the predicted instances in 3D. The middle image shows the same as the left image, but zoomed in. The right-hand image shows the zoomed-in view, but only the predicted results in 3D space.}\label{Fig: Re-projection-graphicalabstract}
\end{figure}

\end{graphicalabstract}

\begin{highlights}
\item Filament during printing/post-processing quality control
\item Data capturing using either Camera, SLS, or TLS
\item Pinhole camera model generation
\item Instance segmentation
\end{highlights}


\end{frontmatter}



\section{Introduction}\label{intro}
Recently, the construction industry has been moving towards fully digitised and automated production \cite{Wolfs2024}. In traditional construction, concrete objects are generally manufactured using moulds \cite{PAOLINI_2019, Doerrie2025}. However, additive manufacturing in construction (AMC) introduces a new perspective on how objects can be created. Toolpaths are generated directly from the 3D model, enabling project-specific production instead of uniform mass production \cite{Doerrie2025}. As a result, additive manufacturing (AM) delivers greater design freedom and mass customization with near-constant manufacturing complexity, production flexibility, improving material efficiency, supports topology-optimization and sustainability \cite{HAGER_2016, GHAFFAR2018, PAOLINI_2019, Cabibihan_2023, Thiel_2024, Messmer_2024}. Despite the aforementioned advantages of AM, proper quality control (QC) is required to ensure the geometric consistency of these somewhat free-form objects. QC is essential during printing and later to identify any defects that might have occurred during the different stages of the printing cycle \cite{Mawas2025_Reviewpaper}.

Two primary techniques for large-scale additive manufacturing in 3D concrete printing (3DCP) are extrusion-based (Contour Crafting--~CC) and shotcrete 3D printing (SC3DP). In both techniques, the concrete material is printed in long filaments (strands) \cite{Dressler2020}. Nevertheless, other additive manufacturing techniques exist, such as slipforming and binder jetting. More details can be found in \cite{Wangler2016}. However, this paper will focus exclusively on the two primary techniques for 3DCP. 

Contour Crafting (CC) is an extrusion-based 3D concrete printing method in which material is deposited layer by layer through a digitally controlled nozzle. The fresh concrete retains its shape upon deposition and becomes self-supporting immediately. In contrast, shotcrete 3D printing (SC3DP) builds components by spraying material layer by layer. Compressed air at the nozzle accelerates the jet, improving layer compaction and interlayer bond strength \cite{KLOFT2020}. 

The application of a new concrete layer is crucially dependent on the time. Thus, any deviation in timing can result in undesirable material deformation. In the worst case, this can lead to the collapse of the printed structure, material displacement, or poor adhesion between layers \cite{Lindemann2019}. Thus, to ensure a smooth flow of concrete without interruptions, additional stress is required, but pinpointing the exact moment for applying this stress is challenging \cite{Senthilnathan2022}. Similarly, the authors in \cite{Mechtcherine2022} investigated layer bonding in extrusion-based processes. The layer bonding is concluded to be affected by mechanical properties and environmental factors. For example, water loss greatly influences filament bonding, thereby impacting the performance of printed objects in terms of their mechanical characteristics and durability. This is due to the high surface-to-volume ratio of printed objects, the lack of protective formwork, and the difficulties involved in curing while printing. These factors increase the risk of water evaporation, changes to the microstructure, shrinkage, and cracking. For more in-depth information on these concepts, readers are encouraged to explore the research conducted in \cite{Quah2023}.

In the context of 3DCP, it is essential to consider a range of processing parameters. These parameters are contingent upon the specific 3D printing technique employed, with notable distinctions between extrusion-based and SC3DP methods. When comparing the two, the extrusion-based method diverges from SC3DP in several key aspects. Notably, extrusion-based printing lacks air pressure at the nozzle, operates with zero compressed air volume, and maintains a fixed nozzle distance, which aligns with the filament's height \cite{KLOFT2020}. 

Furthermore, the precision of printed components is affected by various factors for the SC3DP method, including air volume flow rate, air pressure, accelerator dosage, nozzle parameters (distance, velocity, layer spacing, application angle), and the time between layer applications \cite{Lindemann2019}. Both the deposition rate and the application speed show a nearly linear correlation with layer thickness. Conversely, spraying distance is linearly related to layer width. Meanwhile, air flow significantly impacts material distribution, layer geometry, and rebound. Although this relationship is non-linear but can be treated as a constant during offline process planning. Similarly, in extrusion-based processes, the surface profile of a printed concrete element is influenced by several variables, including nozzle shape and height, filament shape, print path, material flow rate, material rheology (influenced by mixture design), process time, and volume changes during setting and hardening \cite{Nair2022}. Hence, evaluating filament shapes allows direct control over processing parameters \cite{Slepicka2024}.

In this paper, a deep learning approach for filament geometry segmentation is proposed, applicable to both extrusion-based and SC3DP methods. The proposed method is capable of handling both 2D and 3D sensor data, which can be produced from a variety of devices, including a camera, a structured light scanner (SLS), or a terrestrial laser scanner (TLS). Moreover, the method is suitable for different materials, such as concrete and clay. In addition, it can be used for online and post-printing quality control applications. The method is also applicable to fresh and cured state materials.

The remainder of this paper is organised as follows: Section~\ref{SOTA} presents the state-of-the-art on filament quality control for 3DCP for extrusion and SC3DP techniques. Then section~\ref{Methodology} provides a comprehensive overview of the methodological approach employed in this study. Section \ref{Results} presents the results of the proposed method with different data sources, namely, camera-based, SLS, and TLS. Finally, section~\ref{Conclusion} concludes with a discussion of the proposed filament segmentation approach and future work.

\section{Related work}\label{SOTA}
In the literature, researchers have approached the monitoring of the 3D printing process from various perspectives, utilising different sensors and methodologies. Nevertheless, the goal of all approaches is to maintain the width and height of the filaments according to their as-designed model. Consequently, this study will focus on methods based on vision sensors. These have shown promising potential for the inspection of filament geometry, compared to other sensing techniques \cite{KAZEMIAN2021}.

Different researchers use a laser profiler sensor to inspect the shape of printed filaments \cite{Lindemann2019, Lachmayer2022, Lachmayer-VolumeFlow, LACHMAYER2023-ModellingInfluence, Lachmayer-SpatialControl, Jhun2024, VERSTEEGE2025}. The laser profiler offers highly accurate data capture over other sensors, such as depth images \cite{Lindemann2019}. The authors in \cite{VERSTEEGE2025} utilised two Laser profiler sensors for online QC and near post-printing QC for extrusion-based printing. The method relies on creating rectangular cross-sections to monitor the filaments surface and geometry. Similarly, a laser profiler is utilised in \cite{LACHMAYER2023-ModellingInfluence}, for controlling the filament shape for SC3DP. These studies focus on online QC using a laser profiler. However, since the laser profiler is a 1D sensor, it is considered inefficient and time-consuming for capturing entire objects for post-printing quality QC.

In extrusion-based printing, cameras have also attracted the attention of many researchers due to their spatial level of detail, speed, and adequate shape representation \cite{KAZEMIAN2021, Yang2022, Yang2024, RILLGARCIA2022, Barjuei2022, Davtalab2022, Silva2024, Cui2025}. The study in \cite{Cui2025} used a camera sensor to extract filament delineation with the YOLO-v5 model for instance segmentation in extrusion-based 3DCP. The authors observed the layer height as well as the layer angle and connected the observation to the rheological properties for further refinement of the results. Similarly, in \cite{RILLGARCIA2022}, the authors deployed a camera to detect the height of the filament using a deep learning model using a U-Net architecture. In \cite{Silva2024}, the authors developed a real-time machine learning model based on XGBoost to predict the motion speed of the extrusion nozzle in order to control the width of the filament. The model requires a variety of different parameters, including the filament's width. Therefore, a camera sensor is deployed to give a real-time reading of the width of the filament.

In addition to artificial intelligence models, researchers investigated traditional computer vision techniques. The study in\cite{Davtalab2022} delineated the height of the filament geometry. Their method relies on using a Canny edge detection algorithm, followed by a Hough transform algorithm for line extraction. Similarly, \cite{Barjuei2022} deployed a camera for width filament observation in order to control the velocity of the nozzle automatically. The width of the filament is extracted through a traditional computer vision approach, specifically by extracting the contour edges of the top-view surface of the observed filament. Despite the high level of adaptation and accuracy of cameras, as well as their ease of use, Cameras are 2D passive sensors that lack depth information.

In \cite{Villacres_2021}, filament segmentation from 3D point cloud data (PCD) is performed using a k-means clustering algorithm. The authors used a variety of low-cost sensors, namely: RGB-D, RGB-Lidar, and Lidar, to evaluate the height of the layers for deformation analysis. For SC3DP, a highly accurate sensor, such as TLS, is utilised for filament extraction \cite{Mawas2023}. The authors computed the Cloud-to-Mesh (C2M) distance between the point cloud and the as-designed model. Then, the 3D points were projected onto a Y-Z 2D plane. The projected data was then converted into 2D images that included the C2M color distance to extract the contours of the filaments using traditional computer vision techniques. However, this method is only suitable for projecting onto planar surfaces where the object is simple and parallel to one of the main axes.

For clay extrusion-based 3D printing, the author used an SLS to capture 3D data \cite{WI_Clay_SLS}. The author manually generated a variety of profiles for filament analysis. Similarly, for 3DCP specifically for extrusion-based printing, an SLS sensor to capture data is utilized in \cite{Mendricky_Keller_2023_SLSData}. The authors analysed filament deformation using a digital calliper.

The literature review shows that camera and laser profilers are the most commonly used sensors for filament quality control. However, additional research is needed towards a comprehensive approach that combines post-printing and online QC methods, particularly regarding the use of suitable sensors, such as SLS and TLS. 

The research aim is to provide an adaptable filament segmentation from various data sources, including cameras, TLS, and SLS. Thus, sensor independence and applicability for online and post-printing quality control should be reached. An additional goal is to develop a method that can be used for both: SC3DP and extrusion-based printing techniques. Additionally, it should be suitable for various materials, including concrete and clay. Lastly, the method should be used with materials in either a fresh or cured state.

\section{Methodology}\label{Methodology}
\begin{figure}[h]
\centering
\includegraphics[width=1\textwidth]{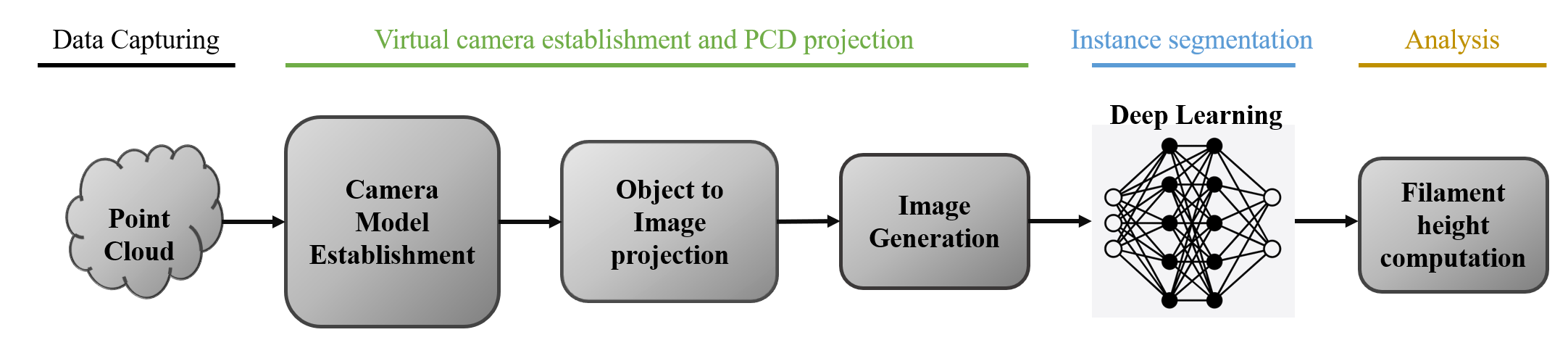}
\caption{Workflow of the proposed method.}\label{Fig: Method_Workflow}
\end{figure}

The proposed method (see Fig.~\ref{Fig: Method_Workflow}) involves establishing a virtual camera (VC) in 3D space to generate an image from a point cloud at a desired pose. The recorded color information has a different meaning based on the sensor used. If an image block, followed by structure from motion techniques and dense image matching, is used, the generated image from the VC would have RGB. However, for TLS usage, the color information is the recorded intensity values of the backscattered signals. Lastly, for SLS, the data is typically a mesh. Thereby, the area of interest is subsampled to generate a PCD. Afterwards, a plane is fitted onto the data, and a cloud-to-mesh (C2M) distance is computed to create a signed distance color for each point.

Consequently, a colorized image is obtained by projecting the point cloud onto the defined VC. Thus, the proposed methodology would be dependent solely on PCD as an alternative to the capturing sensor. Subsequently, an instance segmentation model is employed to extract the filaments from the generated image. The deep learning method is based on Yolo-v11 from Ultralaytics \cite{Ultralytics}. Furthermore, each segmented filament is then processed further to compute the height of the filament by the distance transform algorithm (DT) (cf. Sec.~\ref{Filament heigh}) \cite{DistanceTransform_PAGLIERONI1992, Maurer_DT_2003}.

The advantage of this method is that it not only works on 2D images but also has the benefit of dealing with 3D data. Thus, having segmented instances in 3D space is possible. Since the Cartesian coordinates from the point cloud are already known and being recorded, a backward projection is possible. Moreover, the model segmentation can be used not only for post-processing but also for online QC. Additionally, the model is trained on SC3DP data as well as extrusion-based printing methods.

\subsection{Data Capturing}
The process starts with data acquisition of the printed object, as shown in Fig~\ref{Fig: Method_Workflow}. Point clouds can be obtained from a variety of sensors, such as TLS and SLS. Additionally, photogrammetry methodology can be used to compute PCD through 2D images. For a deeper understanding of data acquisition from a variety of different sensors in terms of methods, and sensor descriptions, the reader is referred to the following paper \cite{Mawas2025_Reviewpaper}. Once the point cloud is acquired, post-processing for denoising, cleaning, and filtering is performed as necessary.

Nevertheless, the process of aligning the data with the robot coordinate system, as well as aligning it with the different stations in the case of the TLS sensor, is described in \cite{Mawas2022_Direct_Coregistration}. The registration is accomplished by utilizing target-based and plane-based registration.

\subsection{Virtual Camera Model Establishment and PCD projection}
\begin{figure}[htb]
\centering
\includegraphics[width=0.85\textwidth]{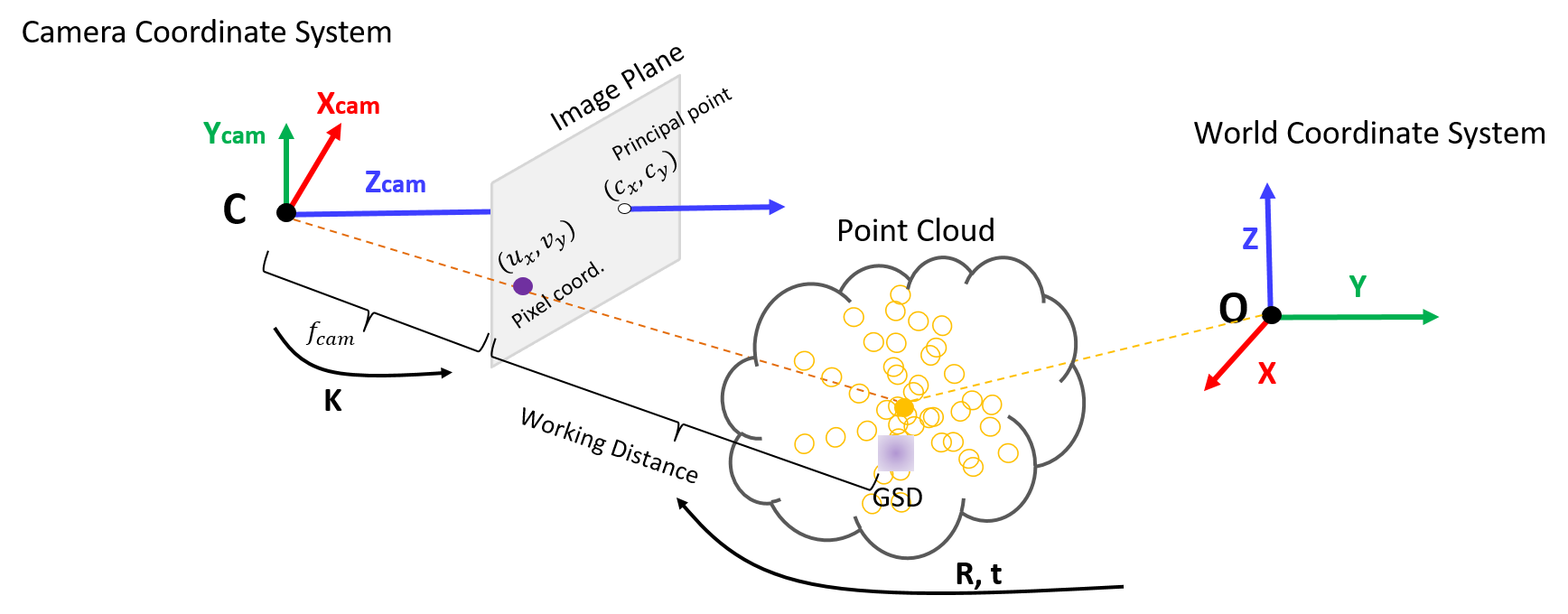}
\caption{Euclidean transformation between the world and camera coordinate frames, Adopted from \citep[p.~156]{Hartley_Zisserman_2004}. K: Intrinsic matrix, R: Rotation matrix, t: Transformation vector, and GSD: ground sampling distance.}\label{projection}
\end{figure}

The virtual camera model is established to project the point cloud onto an image plane to enable a subsequent image-based segmentation. Fig.~\ref{projection} demonstrates the steps and matrices needed to perform the projection of a PCD into the image plane. The required matrices and parameters to be defined to perform the projection are: K\textemdash Intrinsic matrix, R\textemdash Rotation matrix, t\textemdash Transformation vector, and GSD\textemdash ground sampling distance.

\noindent
The intrinsic matrix of the camera is defined by its focal length and the position of its principal point. where \(f_x\) and \(f_y\) are the focal lengths along the \(x\)- and \(y\)-axes, respectively, and \(c_x\) and \(c_y\) denote the principal point coordinates in the image plane. For simplicity, it is assumed that $f_x = f_y$.

In a common convention, for euler angles, a 3D rotation matrix \(R\) can be composed by sequentially rotating about the \(z\)-axis by \(\psi\) angle, then about the \(y\)-axis by \(\phi\) angle, and finally about the \(x\)-axis by \(\theta\) angle. In addition, the 3D translation vector is defined as a vector with three coordinates on the $x-,~y-,~and~z-axes$.

\begin{figure}[ht]
\centering
\includegraphics[width=1\textwidth]{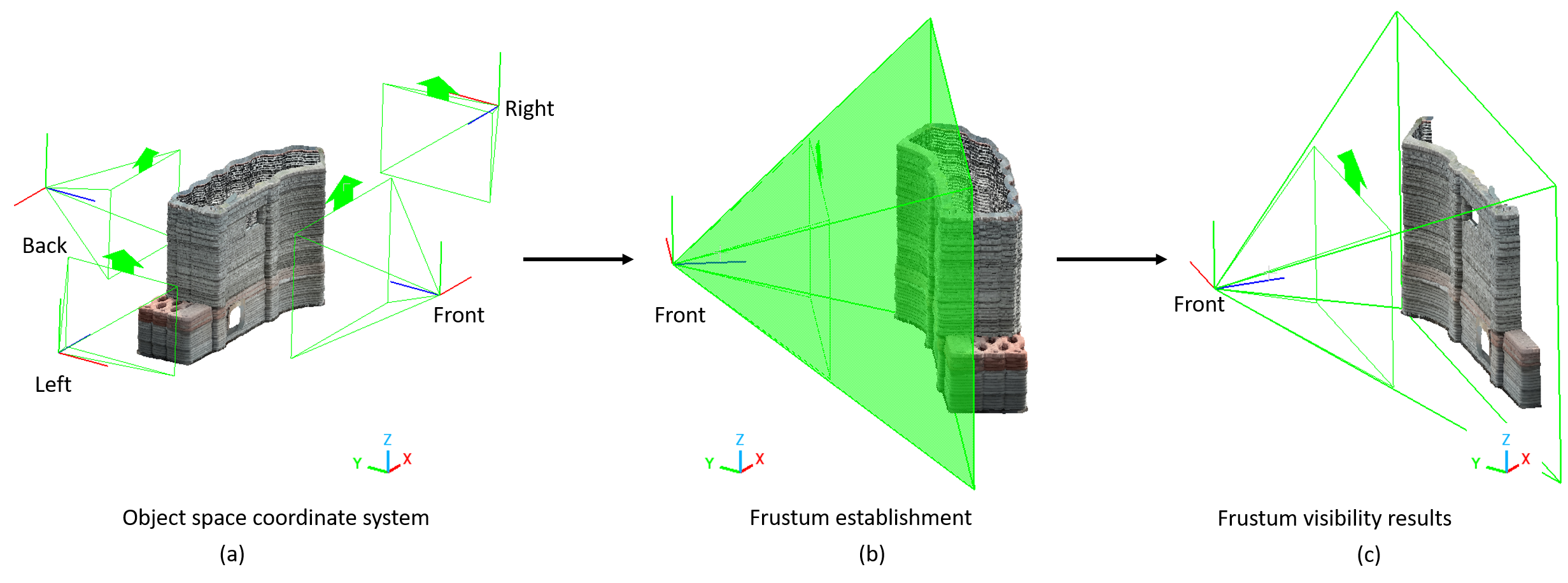}
\caption{Different VC orientation scenarios: (a) predefined four sides, where the camera sensor plane is parallel to one side of the bounding box sides. (b) Frustum Establishment from a VC camera. (c) The points inside the frustum are only considered.}\label{Camera_Orientation_undefined_position}
\end{figure}

The last remaining parameter, namely the ground sampling distance (GSD), should be defined for a proper level of details (LoD) capturing of the region of interest.

\begin{equation}\label{eq:GSD}
\mathrm{GSD} = \frac{D \times \text{pixel size}}{f}
\end{equation}
\noindent
where:
\begin{align*}
    \mathrm{GSD}      & : \text{Ground Sampling Distance} \, [\mathrm{m/pixel}] \\
    D                 & : \text{Camera to object distance, or working distance} \, [\mathrm{m}] \\
    \text{pixel size} & : \text{Physical size of one pixel} \, [\mathrm{mm}] \\
    f                 & : \text{Focal length of the camera} \, [\mathrm{mm}]
\end{align*}

It can be concluded from eq.\ref{eq:GSD}, that the GSD can be controlled by the working distance of the camera. The value of GSD is determined by considering Shannon's sampling theorem to properly reconstruct the smallest details, as proposed in \cite{Shannon1949}. According to the theory, the GSD should be less than half the height of the area between two adjacent filaments. This is the area where adjacent filaments need to be realized in order to distinguish between different filament topologies. After determining the proper distance between two adjacent filaments (in this case $\mathrm{GSD}\le1\,\mathrm{mm}$), the working distance can be computed according to the eq.~\ref{eq:GSD}. Finally, to position the VC in 3D space, the camera's line of sight must be defined. Different scenarios can be realised, namely: (i) predefined position and (ii) known sensor position, as covered in the subsequent paragraphs.

\begin{figure}[htb]
\centering
\includegraphics[width=0.8\textwidth]{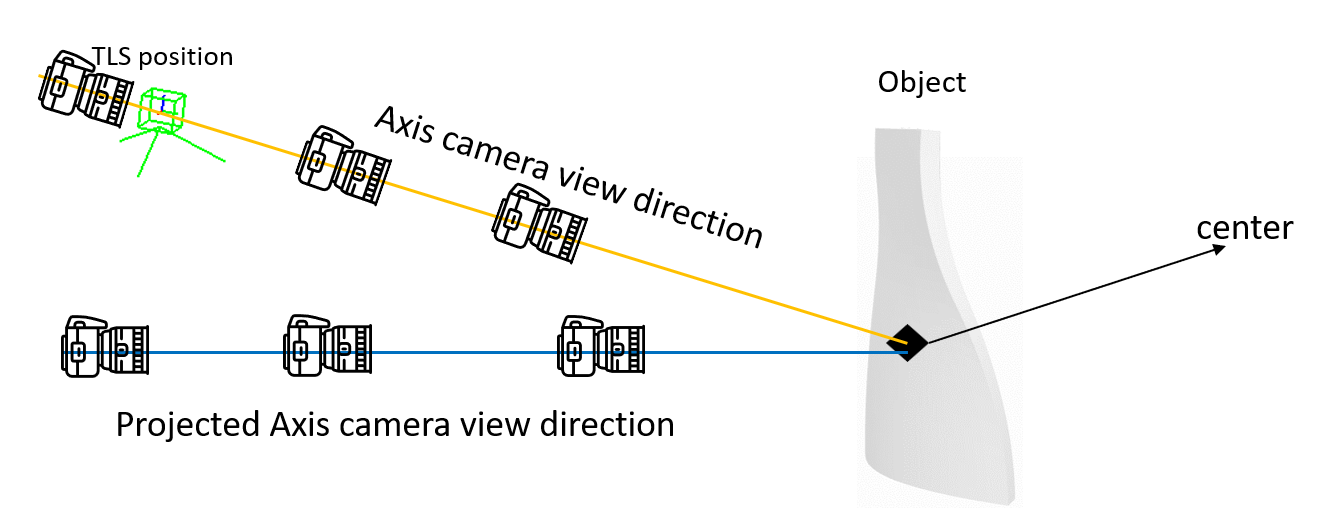}
\caption{VC view direction axes—direct (gold) vs. horizontal-projected (blue) through the point-cloud centroid. Camera placements along these axes to achieve the required working distance and GSD.}\label{VC_Sensor_position}
\end{figure}

\textbf{(i) Predefined position (PP):}
The VC is defined so that the image plane of the camera is parallel to one of the sides of the bounding box (Bounding box side — BBS). Furthermore, the camera's line of sight is perpendicular to the plane of the BBS and intersects it at its center (see Fig.~\ref{Camera_Orientation_undefined_position}). This approach is advantageous for capturing simple, non-complex objects for which a predefined orientation is sufficient. As illustrated in Fig.~\ref{Camera_Orientation_undefined_position}a, various predefined camera positions surrounding a PCD from a photogrammetric approach are demonstrated. Consequently, a frustum is created from the front camera to delineate the front facade (Fig.~\ref{Camera_Orientation_undefined_position}b). The final stage of this process is illustrated in Fig.~\ref{Camera_Orientation_undefined_position}c, which displays the clipped points.

In order to avoid the projection of actually hidden data into the VC, e.g. when the object is captured from all sides, the point cloud should be segmented before the overall process is initiated. As a result, once the VC has been established based on the predefined orientation and computed working distance, a frustum is created to remove unwanted points that lie outside it. However, this step is performed manually. The points inside the frustum are then back-projected into the camera, retaining their colour information (RGB, intensity, etc.).

\textbf{(ii) Known sensor position (KSP):}
The second approach is valid when the sensor's position at the data capture stage is known. This is particularly useful for an automated approach in which only the area of interest (AoI) is captured rather than the entire scene. This can usually be achieved using TLS, where the sensor position can be obtained from the data header. Defining the VC’s view direction in this way helps to automate the image generation process, as it eliminates the need for manual adjustments by the user.

In addition, generating an image from a direction other than the original viewing direction of the capturing sensor introduces many artefacts and noise, namely: self-occlusion and spatial aliasing. Self-occlusion occurs because the sensor captures the surface from its line of sight, and anything behind that surface is never actually captured. Therefore, changing the direction of the line of sight results in holes and range shadowing. Spatial aliasing, especially around edges, is caused by undersampling artefacts, which can be referred to as viewpoint aliasing.

Fig.~\ref{VC_Sensor_position} shows two approaches of defining the VC view direction axis. One approach is to define the axis that crosses the position of the captured sensor and the centroid of the PCD's AoI. The alternative approach involves projecting the aforementioned view direction axis onto a horizontal plane that crosses the centroid of the PCD. Finally, the camera can be placed on either axis and moved away from the centroid to match the calculated working distance, thereby meeting the GSD requirement.

\begin{figure*}[ht]
\centering
\includegraphics[width=0.9\textwidth]{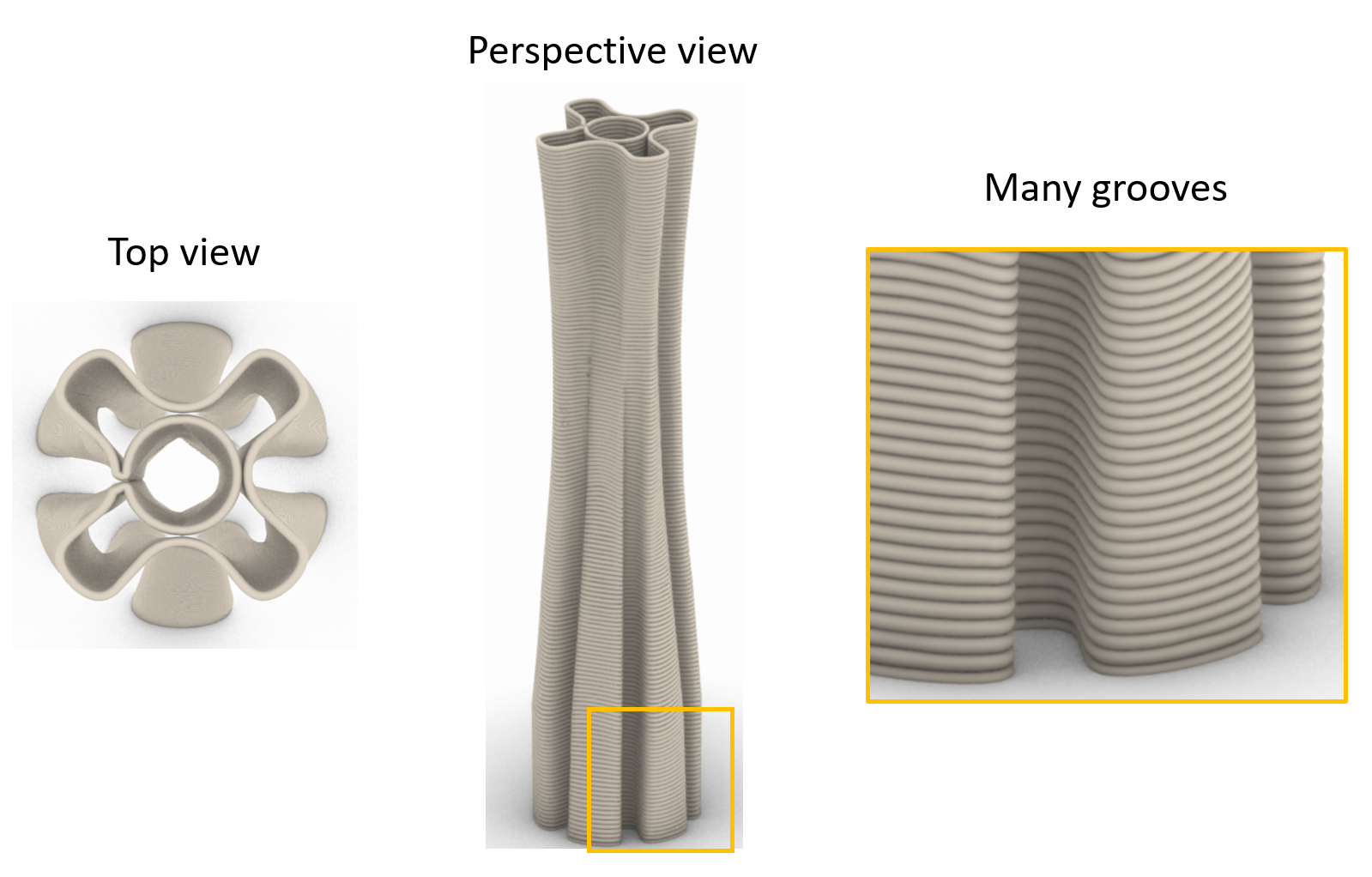}
\caption{Complex shape - shown in top and perspective views, contains many grooves which can make it difficult to define the VC view direction.}\label{Shape_Complexity}
\end{figure*}

The \textit{known sensor position (KSP)} approach offers a key advantage over the \textit{predefined position (PP)} approach, especially for complex geometries. As the sensor positions have already been determined, either by the surveyor in the field or via a pre-planned algorithm, the KSP method uses the actual camera poses from the data capture. This approach enhances time efficiency and data quality by eliminating redundant planning and minimizing spatial aliasing, particularly for complex objects, as illustrated in Fig.~\ref{Shape_Complexity}.

\subsection{Image generation and instance segmentation}\label{Sliding_window}
Since an image can cover an entire side of an object and images result in a variety of different resolutions, a sliding window approach is adapted to ensure that the images fed into the model match the model requirements and to avoid resizing steps, thus maintaining the required resolution.

The model used for instance segmentation is based on You-Only-Look-Once Yolo-v11 \cite{Ultralytics}. The model is fine-tuned on the data captured (cf. Sec.~\ref{datasets}), which were taken from the aforementioned sensors after generating the back-projected images.

\subsection{Filament height}\label{Filament heigh}
The geometry of the filament is influenced by a variety of variables, such as nozzle velocity, flow rate, and material properties, to name a few (cf. Sec.~\ref{intro}). Direct control of this combination of parameters can be achieved by extracting the thickness (height) of the printed filament. The thickness is extracted from the predicted filament mask using a distance transform operation \cite{DistanceTransform_PAGLIERONI1992, Maurer_DT_2003}. The thickness of the predicted filament is obtained by calculating the distance of each pixel in the image to the nearest background pixel (the edge of the layer) using a distance transform algorithm. In addition, the maximum peak at each vertical column is taken from the binary distance result to construct a profile of the thickness across the entire filament body (cf. Sec.~\ref{Filament height analysis}).

\section{Experiment}
\subsection{Available datasets}\label{datasets}
\begin{figure*}[t]
\centering
\includegraphics[width=1\textwidth]{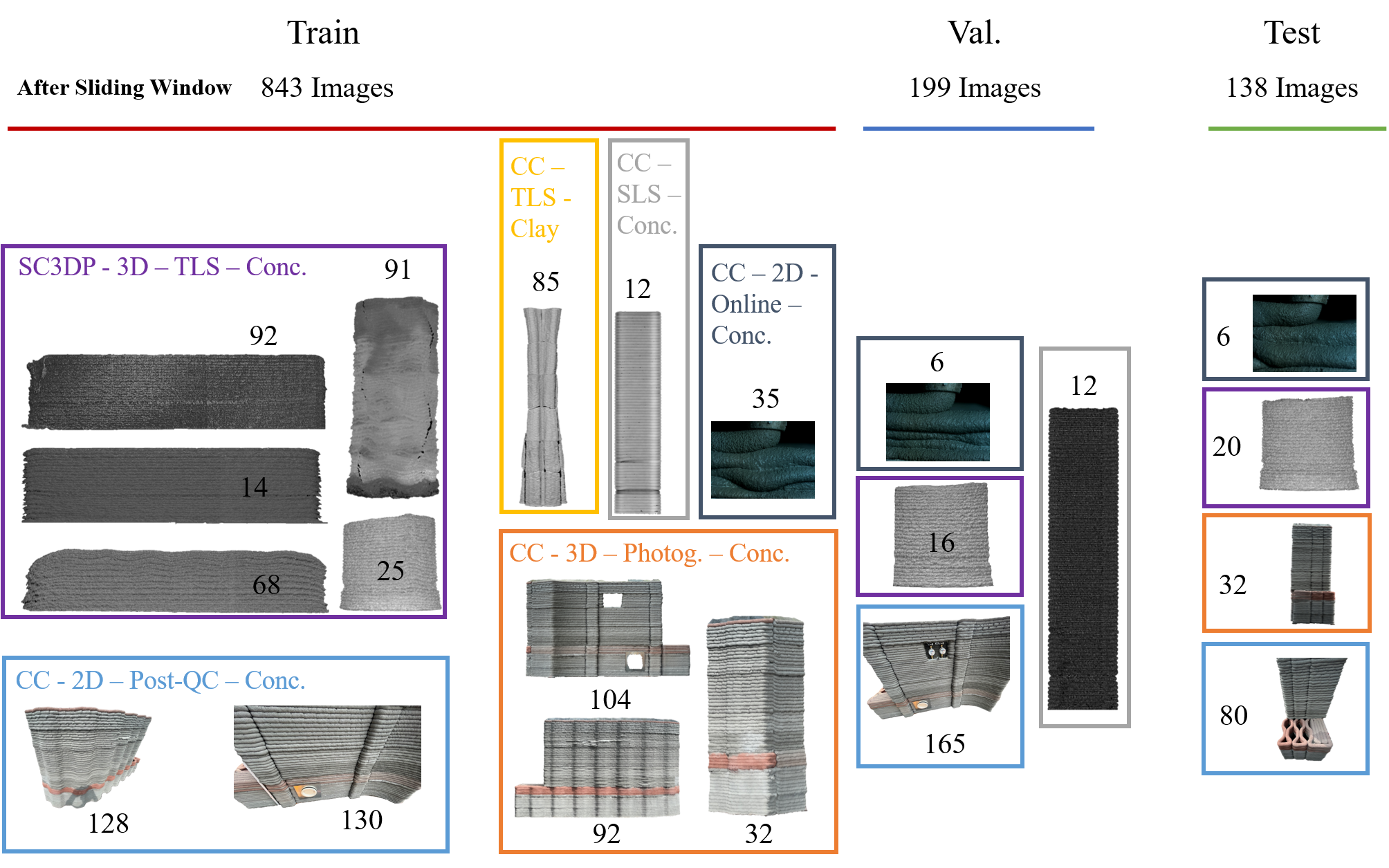}
\caption{Dataset used for instance segmentation deep learning model. The data is categorized based on their printing method, sensor used, and material state. The number beside each object refers to the number of images resulting from the sliding window method.}\label{Fig: Data_Summary}
\end{figure*}

The data used in this study are diverse, not only in terms of sensor sources but also in terms of the materials and printing technologies (see Table~\ref{tab:data_sources}). The 2D real-time monitoring images used in this study were obtained from Rill-García et al. (2022) \cite{RILLGARCIA2022}, while the 3D point-cloud data acquired via structured-light scanning (SLS) were sourced from Mendricky and Keller (2023) \cite{Mendricky_Keller_2023_SLSData}. TLS datasets for SC3DP encompass multiple objects exhibiting a range of geometric complexities. For further insights into the dataset used, please refer to Fig.~\ref{Fig: Data_Summary}. Additionally, further explanations about the data can be found in Sec.~\ref{training_and_validation}.

\begin{table}[htbp]
  \centering
  \footnotesize
  \caption{Data from different sensors and materials using 3D printing deposition technology.}
  \label{tab:data_sources}
  \begin{tabularx}{\textwidth}{l X l}
    \toprule
    \textbf{Printing Technology} & \textbf{Sensor / Data Type} & \textbf{Material} \\
    \midrule
    \multirow{5}{*}{Extrusion‐based}
      & RGB / Images (real-time monitoring camera)~\cite{RILLGARCIA2022}           & Concrete \\
      & RGB / Images (2D distorted perspective) & Concrete \\
      & RGB / Point Cloud (via photogrammetry)       & Concrete \\
      & SLS / Point Cloud ~\cite{Mendricky_Keller_2023_SLSData}                         & Concrete \\
      & TLS / Point Cloud                   & Clay     \\
    \midrule
    SC3DP
      & TLS / Point Cloud (Several objects)                   & Concrete        \\
    \bottomrule
  \end{tabularx}
\end{table}

The dataset contains a diverse collection of images and point clouds of 3D-printed concrete objects. As illustrated in Fig.~\ref{Fig: Data_Summary} data sources include 2D images with perspective distortion of CC 3D-printed structures (in light blue), fresh extrusion-based images from Rill-García et al. (2022) \cite{RILLGARCIA2022} (in dark blue), photogrammetry-based 2D images (in orange) , and 3D point clouds acquired via SLS reported in \cite{Mendricky_Keller_2023_SLSData} (in gray).

Additionally, TLS datasets include both concrete (in violet) and clay samples (in yellow) of varying geometries. While 2D images are ready for direct deployment by deep learning models, point cloud data undergoes a virtual camera projection process to convert spatial data into 2D images, as explained in Fig.~\ref{Fig: Method_Workflow}.

To optimise image resolution for segmentation, the $\mathrm{GSD}\le1\,\mathrm{mm}$ is set in accordance with Shannon’s sampling theorem, ensuring that details below half the inter-filament height are captured. All images are then standardized to 512×512 pixels using a sliding window approach (cf. Sec~\ref{Sliding_window}) to accommodate variable object sizes. The total number of images resulting from the sliding window is as follows: 843 images for training, 199 images for the validation dataset, and 138 images for the test dataset.

\subsection{Results}\label{Results}
\subsubsection{Training and validation}\label{training_and_validation}

Subsequent to the preparation of the datasets (cf. Sec.~\ref{datasets}), a Yolo11s-seg model (small version) is fine-tuned to perform instance segmentation of filaments in 3DCP. Input images are $512 \times 512~[pixels]$ tiles resampled to a GSD of $\le 1~mm$. The network is initialized with weights that have been pre-trained on the COCO dataset, and its mask head is adapted to align with the filament annotations provided. To ensure the effective training and evaluation of the models, a training and validation dataset was utilized (see. Fig.~\ref{Fig: Data_Summary}).

In the course of the training, a batch size of 32 was employed in conjunction with early stopping, and two distinct optimizers, namely AdamW and SGD \cite{AdamW, SGD}, were utilized. Furthermore, several data augmentations were utilized during training, including random left-right flipping, random erasing, mosaic augmentation, HSV-based color jittering, random translation, and random scaling. Training was conducted on Tesla P100-SXM2-16GB GPU, utilising automatic mixed precision (AMP) acceleration.

\begin{table}[ht]
\centering
\caption{Precision, Recall and mAP values at different $IoU$ thresholds for bounding box and mask reported on the validation dataset. Highest values per metric are highlighted in bold.}
\label{tab:yolov11-results_training_validation}
\resizebox{\textwidth}{!}{
\begin{tabular}{lcccccccc}
\toprule
Optimizer & \multicolumn{4}{c}{Box} & \multicolumn{4}{c}{Mask} \\
\cmidrule(lr){2-5} \cmidrule(lr){6-9}
& Precision & Recall & mAP$^{0.50}$ & mAP$^{0.50:0.95}$ 
  & Precision & Recall & mAP$^{0.50}$ & mAP$^{0.50:0.95}$ \\
\midrule
AdamW & 0.596 & 0.540 & 0.468 & 0.255 & \textbf{0.701} & \textbf{0.549} & \textbf{0.518} & \textbf{0.289} \\
SGD   & 0.596 & \textbf{0.544} & \textbf{0.497} & \textbf{0.269} & 0.632 & 0.473 & 0.454 & 0.248 \\
\bottomrule
\end{tabular}
}
\end{table}

Tab.~\ref{tab:yolov11-results_training_validation} shows different behaviour with respect to the results for bounding boxes and segmentation. The AdamW optimizer outperformed SGD for mask segmentation. While SGD revealed better performance for detecting the filaments. Additionally, the result data revealed the highest F1-score of 0.62 for filament detection at a confidence level of 0.5.

\subsubsection{Testing}
To assess the efficacy of the trained models, they have been deployed on the test dataset, as illustrated in Fig.~\ref{Fig: Data_Summary}. The test dataset consists of 2D images of fresh extruded material. There are also 2D images of a cured material with a distorted perspective. Additionally, there are VC renderings of an SC3DP object captured with TLS, as well as a point cloud from a photogrammetry approach of an extruded-based printing object.

The processing time for each stage is shown in Tab.\ref{tab:processing-times-comparison}. As indicated, the end-to-end processing time per image was approximately 13 milliseconds, equivalent to around 76 frames per second (FPS), which is usually suitable for online QC applications.

\begin{table}[ht]
  \centering
  \caption{Average processing times per image (ms) by the trained model.}
  \label{tab:processing-times-comparison}
  \begin{tabular}{l r r}
    \hline
    Stage            & (ms) \\
    \hline
    Pre-processing   & 0.283     \\
    Model inference  & 5.987      \\
    Post-processing  & 6.794      \\
    \hline
    Total time            & 13.06     \\
    \hline
    FPS (frames per second)            & 76.54     \\
    \hline
  \end{tabular}
\end{table}

Tab.~\ref{tab:performance-metrics-combined} shows the different performance of the two models studied with two different optimizers, SGD and AdamW, for detection boxes and segmentation. Additionally, the overall fitness metric shows that the SGD model performs better. The fitness metric is a weighted combination of box and segmentation metrics for $mAP_{50}~and~mAP_{50:95}$ \cite{Ultralytics}.

\begin{table}[ht]
  \centering
  \caption{Precision, recall, and mAP metrics for bounding box (B) and mask (M) predictions, obtained using AdamW and SGD optimizers on the test dataset. Highest values per metric are highlighted in bold.}
  \label{tab:performance-metrics-combined}
  \resizebox{\textwidth}{!}{
    \begin{tabular}{lcccccccc}
        \toprule
        \multirow{2}{*}{Optimizer} 
        & \multicolumn{4}{c}{Box (B)} 
        & \multicolumn{4}{c}{Mask (M)} \\
        \cmidrule(lr){2-5} \cmidrule(lr){6-9}
        & Precision & Recall & mAP$_{50}$ & mAP$_{50:95}$ 
          & Precision & Recall & mAP$_{50}$ & mAP$_{50:95}$ \\
        \midrule
        AdamW & 0.5722  & 0.4413 & 0.4436 & 0.2193  & \textbf{0.6184} & 0.4279 & 0.4361 & 0.2329 \\
        SGD   & \textbf{0.5898} & \textbf{0.5041} & \textbf{0.5332} & \textbf{0.2742}  & 0.5873 & \textbf{0.4408} & \textbf{0.4541} & \textbf{0.2568} \\
        \midrule
        \multicolumn{1}{l}{Fitness (overall)} 
          & \multicolumn{4}{c}{0.4949} & \multicolumn{4}{c}{\textbf{0.5766}} \\
        \bottomrule
    \end{tabular}
  }
\end{table}

As indicated in Tab.~\ref{tab:performance-metrics-combined}, evaluation on the test dataset indicates that the model trained with the SGD optimizer demonstrates superior generalisation compared to the AdamW optimizer model, outperforming it in all metrics except precision.

\begin{figure}[htb]
\centering
\includegraphics[width=1\textwidth]{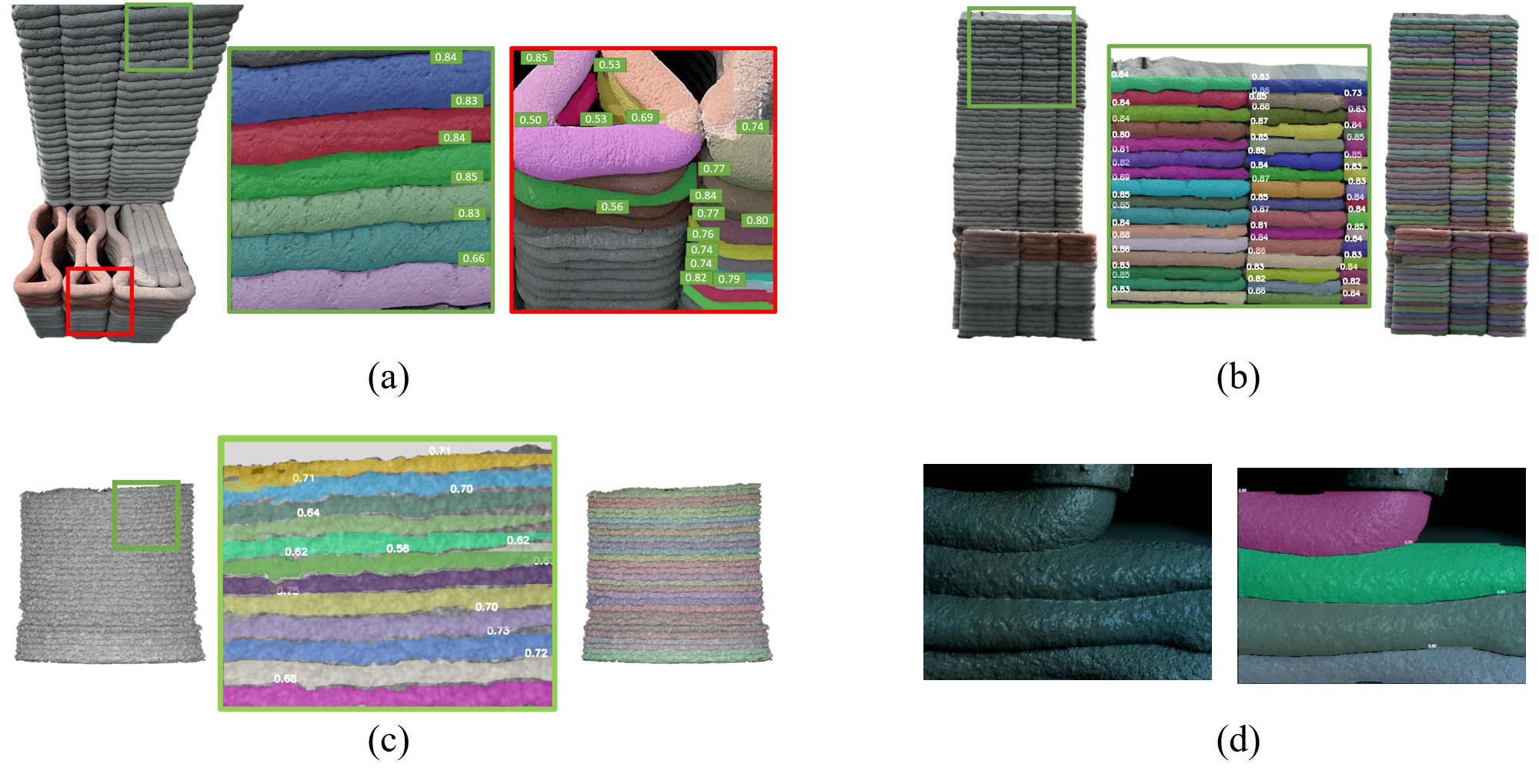}
\caption{SGD-model segmentation results on images after sliding-window processing. The confidence values for each segmented mask are overlaid, and the full-side views are superimposed with the detected masks. (a) 2D extrusion-printed image with perspective distortion. (b) VC render of an extrusion-based from 3D-print point cloud. (c) VC render of an SC3DP object from 3D-print point cloud. (d) An image of freshly extruded concrete material.
}\label{Fig: Results_imposed}
\end{figure}

As shown in Fig.~\ref{Fig: Results_imposed}, the SGD-model produces a mask indicating the filament area superimposed on the input image after the sliding window. The produced masks are used to colourise the results of the different filaments.

The results are based on several cases (see Fig.~\ref{Fig: Results_imposed}), namely: for extrusion base: (i) RGB-images from a traditional camera from real-time monitoring (Fig.~\ref{Fig: Results_imposed}d), (ii) RGB cameras with and without distortion perspective (from photogrammetry)  Fig.~\ref{Fig: Results_imposed}a and b, (iii) TLS data (VC) for SC3DP  (Fig.~\ref{Fig: Results_imposed}c).

The results show good delineation with high confidence of the filaments across different printing methods and material states, as well as across different sensors. This is clearly seen in the extrusion-based approach with no perspective distortion, Fig.~\ref{Fig: Results_imposed}b. Despite of the perspective distortion, some have good delineation results, Fig.~\ref{Fig: Results_imposed}a. Also, for fresh state concrete, the model gave a good prediction Fig.~\ref{Fig: Results_imposed}d. Moreover, for SC3DP, where the filament geometry is more distorted than in the extrusion-based technique and the component shape is more complex, with a helical filament geometry rather than a straight line, the model produced good results Fig.~\ref{Fig: Results_imposed}c.

Nevertheless, the results show that further enhancements to the model are required, particularly for images with high perspective distortion and width-surface filament segmentation (top surface) (see Fig.~\ref{Fig: Results_imposed}a). This is due to the small number of training images consisting of such cases. Furthermore, in Fig.~\ref{Fig: Results_imposed}b and c, the sliding windows are reattached to their positions in the grid and then merged. To properly merge the filaments, each filament instance is merged with its neighbouring instance on the basis of an intersection-over-union (IoU) threshold.

\subsubsection{Filament height analysis}\label{Filament height analysis}
\begin{figure}[ht]
\centering
\includegraphics[width=1\textwidth]{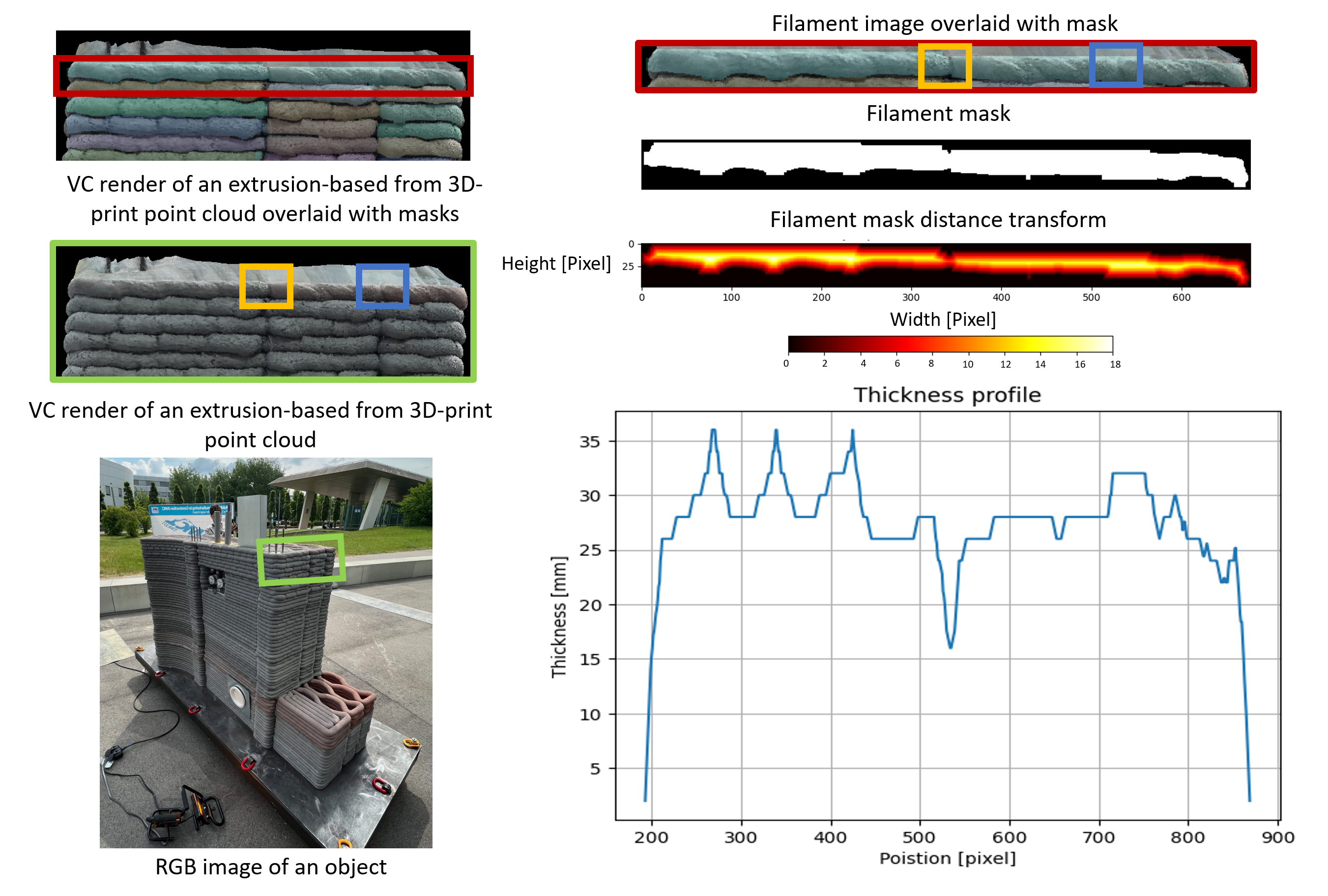}
\caption{Height analysis of a single filament. The images on the left provide an overview of the object with predicted masks from the model. The images on the right show the height analysis of the topmost filament on the right-hand side of the object. The top image displays a single filament overlaid with a predicted mask and a binary mask of a single filament instance. The middle image shows a distance transform, colour-coded according to the distance of each pixel to the nearest edge. The bottom plot presents the filament’s thickness profile.
}\label{Fig: Height analysis}
\end{figure}

After instance segmentation, each instance is analysed further for height analysis by applying a DT algorithm to the binary image. The height analysis for one instance is shown in Fig.~\ref{Fig: Height analysis}. The profile thickness is computed as the mean distance across each column of the DT image. Consequently, the filament thickness is twice the height of the distance transform results. 

As can be seen in Fig.~\ref{Fig: Height analysis}, the yellow-highlighted area shows a sudden drop in thickness. This is because the filament is not continuous on this side. As can be seen in the full image of the object, the filament has a honeycomb shape. This is also where the two materials used to print the object meet. Nevertheless, another separation in the filament geometry can be seen in the blue-highlighted area. However, the absence of any diminution in thickness in this area can be attributed to the fact that the model recognizes it as one continuous filament. Better training is therefore required for such cases to improve model segmentation performance.

For subsequent filaments in extrusion-based 3D concrete printing, the nozzle height at index~$t$ is calculated as the sum of the nozzle height at~$t-1$ and the thickness of the filament. As a result, the measured filament thickness can be compared to the nozzle height specified by the path planning process. In quality control analysis, this relationship can be inverted: the planned nozzle height at~$t-1$ corresponds to the maximum height of the filament at~$t$ along the filament height profile.

\subsubsection{Back-Projection of 2D Segmentations into 3D Space}

\begin{figure}[ht]
\centering
\includegraphics[width=1\textwidth]{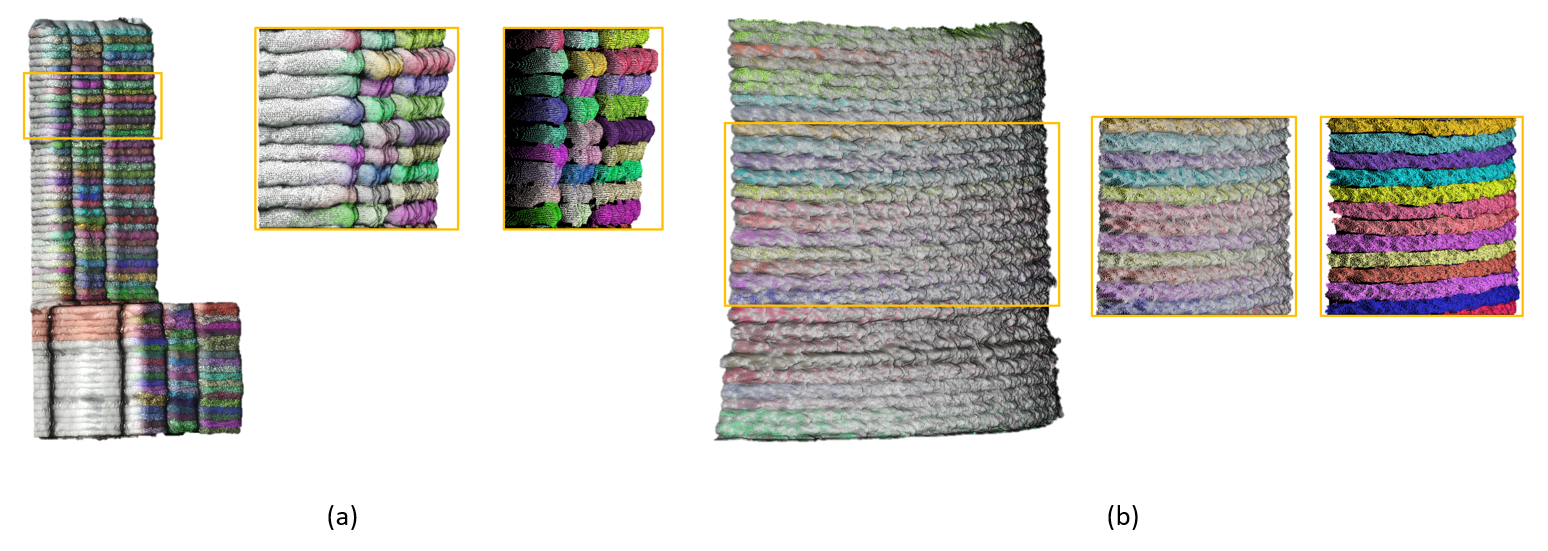}
\caption{The prediction results of the filament instances are projected back into 3D space. (a) Extrusion-based object from Photogrammetry: The results shown are for the data from Fig.~\ref{Fig: Results_imposed}b. (b) SC3DP object from TLS: The results shown for the data from Fig.~\ref{Fig: Results_imposed}c. The left image shows the raw data of the whole object superimposed with the predicted instances in 3D. The middle image shows the same as the left image, but zoomed in. The right-hand image shows the zoomed-in view, but only the predicted results in 3D space.
}\label{Fig: Re-projection}
\end{figure}

Cartesian coordinates are preserved during the point cloud projection procedure. Therefore, it is possible to perform an inverse transformation after instance segmentation. Fig.~\ref{Fig: Re-projection} shows the PCD of the object from Fig.~\ref{Fig: Results_imposed}b\&c.

Nevertheless, closing the feedback loop is important for bridging the two spaces, namely the design space and the physical world. The results from the deep learning model, as well as the QC analysis of the filaments, can be sent back to the BIM/FIM (Building/Fabrication Information Model) to update it on the current state of the design.

\section{Conclusion}\label{Conclusion}
The work presents a robust and flexible quality control (QC) pipeline for filament height analysis in 3D concrete printing (3DCP), addressing both extrusion-based and SC3DP techniques. The proposed methodology is independent of specific sensor hardware because it leverages diverse data capture methods, including traditional RGB imaging, SLS and TLS. The use of virtual camera models allows projection of 3D data into 2D, facilitating instance segmentation via the lightweight YOLOv11 model. The method revealed the model performance accross varying image modalities and filament geometries, including challenging cases such as helical SC3DP filaments and different concrete states, namely fresh-state and cured-state concrete. 

The efficiency of the inference times further supports the application of this approach to both online and post-process QC, which is vital for in-situ monitoring and verification during additive manufacturing workflows. Furthermore, height analysis via distance transform algorithms provides quantitative geometric feedback at the instance level. Consequently, the filament height analysis provides a direct method for controlling the height of the printing nozzle. Additionally, back-projection into 3D space ensures the spatial coherence of predictions for integration into digital twins, BIM or FIM systems. This closed feedback loop connects the gap between physical fabrication and digital design environments, enabling real-time updates and adaptive control of the printing process.

Despite these promising results, limitations were identified, particularly in scenarios involving high perspective distortion. Generally, there is more room for enhancing the model performance and collecting more data is required. Also, the merging process of the different filaments after model deployment requires enhancement. Furthermore, enhancement is needed where the filament geometry is not continuous, to prevent the model from predicting it as one continuous object. In regard to the automation of the process for establishing VC in the context of predefined positions (PP) for photogrammetric data, manual establishment of the frustum for point clipping becomes unnecessary. The hidden points removal algorithm is a viable option for this purpose. As a result, hidden points behind the AoI can be removed. Additionally, rather than colorizing the PCD from SLS with C2M, the depth-based color from the desired position can be tested to automate the process for SLS data. Consequently, a generalizable framework for non-planar surfaces is hereby proposed.

Further analysis of the filaments can be conducted in future, such as examining texture and surface smoothing. Furthermore, filament segmentation on PCD is also definitely worth investigating in the context of 3D data representation. In summary, the proposed pipeline provides a scalable foundation for comprehensive, data-driven, automated quality control in feedback-driven, digital construction technologies for 3DCP.

\vspace{4mm}

\noindent
\textbf{Acknowledgment}: This research is funded by the Deutsche Forschungsgemeinschaft (DFG, German Research Foundation) – TRR 277/2 2024 – Project number 414265976. The authors thank the DFG for the support within the CRC / Transregio 277 - Additive Manufacturing in Construction (Project C06: Integration of Additive Manufacturing into a Cyber-Physical Construction System). In addition, the data acquisition was made possible through a large-scale research equipment investment by DFG, project number 461109100 (gepris.dfg.de/gepris/projekt/461109100)
\\
We would like to express our profound gratitude to Noor Khader from ITE TU Braunschweig and the students of the Additive Manufacturing course for developing the as-designed model in Fig.~\ref{Shape_Complexity}, which was instrumental in creating our figure.
\\
We gratefully acknowledge Radom\'{\i}r Mend\v{r}ick\'{y}
(\texttt{radomir.mendricky@tul.cz})
from the Technical University of Liberec for kindly sharing the scanned data from SLS that was used to train our model.
\\
\noindent
\textbf{Github}: \href{https://github.com/KaramMawas/3DCP_DeepFilament.git}{%
https://github.com/KaramMawas/3DCP\_DeepFilament.git}

\noindent
\textbf{Dataset}: \href{https://doi.org/10.5281/zenodo.17695185}{10.5281/zenodo.17695185}.

\bibliographystyle{elsarticle-num} 
\bibliography{references}

\end{document}